\documentclass[11pt,a4paper]{article}

\usepackage[top=25mm,bottom=20mm,left=25mm,right=20mm]{geometry}
\usepackage[round,sort]{natbib}
\usepackage[draft]{graphicx}

\usepackage{fancyhdr}
\begin{document}
\begin{center}
{\large \bf On the comparison of plans:\\Proposition of an instability measure\\for dynamic machine scheduling}\\[4ex]

{{\bf Martin Josef Geiger}\\[0.5ex]Helmut Schmidt University, Logistics Management Department,\\Holstenhofweg 85, 22041 Hamburg, Germany, \\ {\tt m.j.geiger@hsu-hh.de}}\\[4ex]
\end{center}

{\bf ABSTRACT.} On the basis of an analysis of previous research, we present a generalized approach for measuring the difference of plans with an exemplary application to machine scheduling. Our work is motivated by the need for such measures, which are used in dynamic scheduling and planning situations. In this context, quantitative approaches are needed for the assessment of the robustness and stability of schedules.\\
Obviously, any \lq{}robustness\rq{} or \lq{}stability\rq{} of plans has to be defined w.\,r.\,t.\ the particular situation and the requirements of the human decision maker. Besides the proposition of an instability measure, we therefore discuss possibilities of obtaining meaningful information from the decision maker for the implementation of the introduced approach.\\[2ex]


\section{\label{sec:dynamic:scheduling}Scheduling in dynamic environments}
Planning and scheduling generally considers a set of activities (operations) for which (i) assignments to resources, such as machines, and (ii) definition of starting times must be found, thus defining a {\em schedule} $x$ for a given problem. While the assignment of operations to machines is often given by technical side constraints, the processing order on the machines, and thus the precise starting times of the jobs is determined by the planning procedure/ algorithm. W.\,r.\,t.\ the terminology used in machine scheduling, operations, denoted as $O_{jk}$, are often grouped into \lq{}jobs\rq{} $J_{j} = \{ O_{j1}, \ldots, O_{jo_{j}} \}$, and precedence constraints among the operations belonging to a particular job implement the technical requirements of processing.\\
Besides these general, static considerations, practical problems of scheduling in manufacturing environments are
in most cases of dynamic nature. In these situations, a schedule
needs to be found that is modified during the actual execution,
adapting to the dynamically changing conditions.\\
Changes of relevance to the scheduling are referred to as {\em rescheduling factors} \citep{dutta:1990:article}. Along with these issues come modifications of the relevant data and the characteristics of the problem. Depending on the changes and on the initial schedule $x_{t}$ at time $t$, a revision of $x_{t}$ might be necessary.

Such dynamic adaptations raise stability and robustness issues. Obviously, an impact on the organization has to be expected when making changes to once defined and committed plans. Therefore, quantitative approaches of measuring the difference of plans, and thus the robustness of schedules, are needed for such practical applications.

\paragraph{Rescheduling factors}\hspace{0cm}\\
Changes to the machine environment are the most commonly discussed
factors of dynamic scheduling in the literature. Machine failures
are reported to play a role in dynamic situations \citep{abumaizar:1997:article}. Besides the machines failures due to technical problems, \citet{church:1992:article} report the possibility of operator absenteeism which may lead to an unavailability of the machine.\\
Job related rescheduling factors comprise the arrival of new jobs,
and especially rush jobs \citep{jain:1997:article}. Besides, the cancelation of jobs is another source of dynamics in scheduling.
In the case of defined due date, changes may also address the due
dates \citep{fang:1997:article}.\\
In addition to the basic data of the problem, preferences of the
decision maker and priorities of jobs may change over time, an
example being unexpectedly arriving rush orders which implicitly
reduce the relative importance of known jobs from the previous
planning period \citep{jain:1997:article}.

\paragraph{Classification of solution approaches}\hspace{0cm}\\
Common to all dynamic scheduling concepts is the distinction between
an initial {\em scheduling} and a recurring {\em schedule
adaptation} phase. Two aspects of schedules play an important role when dynamically revising schedules, utility and stability. While the term {\em
utility} refers to the quality of the schedule, expressed by the
objective function(s), the {\em stability} measures the changes of
an initial schedule compared to a revised schedule. The goal of any
rescheduling concept is the maximization of both the achieved
utility as well as the stability. However these two are often of
conflicting nature, they lead to a situation where a tradeoff
between the two aspects has to be made.

The general treatment of dynamic scheduling problems can be broadly
classified into three categories: on-line scheduling,
predictive-reactive and robust scheduling \citep{aytug:2005:article}.\\
{\em On-line} approaches make assignment and sequencing decision on
the job floor in real time \citep{mehta:1999:article}, e.\,g.\ by the
means of dispatching rules \citep{haupt:1989:article}. While this concept is highly flexible, a drawback lies in the missing predictability of
the schedules' performance as the precise production schedule is
only available after the operations have been assigned to the
resources.\\
{\em Predictive-reactive} approaches introduce a
two-step-process by first constructing an initial schedule which is
then to be modified given dynamic changes of the problem. Here, an
important question is the appropriate schedule adaptation to the
changed circumstances, taking into consideration both utility and
stability. Potential changes in the manufacturing environment are
however not anticipated or integrated into the scheduling procedure.\\
Concepts of {\em robust scheduling} try to anticipate possible
changes and sources of disturbances when proposing a schedule for
the problem at hand. While at a first glimpse this concept seems
overall promising, knowledge about possible changes has to be
available or at least sensible assumptions have to be made. One
possibility is the consideration of a range of
scenarios \citep{kouvelis:2000:article}.

\section{\label{sec:stability:of:schedules}Stability of schedules}
\paragraph{Stability measures}\hspace{0cm}\\
Besides the evaluation of schedules with respect to known and
initially defined optimality criteria \citep{tkindt:2002:book},
the {\em stability} of a schedule plays a role in the context of dynamic scheduling. Various stability measures have been proposed in order to express transition between two schedules in a quantitative way. Common to
all is the proposition of a cost function based on the operations
and/or jobs of the problem. Associated with changing starting times
is a negative impact which is penalized in a quantitative way. The
idea behind this can be seen in resulting adaptations that have to
be made in the manufacturing environment, resulting in costs.

Most approaches go back to the early proposal of \citet{wu:1993:article}. Here, the difference between two schedules is computed as the sum of the absolute
differences of starting times of all operations, given in
Expression~\ref{eqn:stability:1}. Stability is therefore independent
from whether operations start earlier or later than initially
planned. The stability $stab$ is expressed depending on the initial
schedule $x$ with starting times $s_{jk}$ and the revised schedule
$x'$ with starting times $s_{jk}'$ of operations $O_{jk}$.

\begin{equation}
\label{eqn:stability:1} stab(x, x') = \sum | s_{jk} - s_{jk}' |
\end{equation}

The stability measure of \citet{wu:1993:article} comes with the implicit assumption, that a discrimination between operations being shifted forward {\em or} backward in time has an equal impact on the stability. It therefore is only applicable in situations in which this assumption holds. In many production situations however, a
difference between delayed operations and earlier executed
operations can be observed. Delaying operations does not necessarily
result in a higher organizational overhead, but it has an obvious
effect on the completion of the jobs, and is in this sense already
captured by other optimality criteria. An earlier execution of
operations however often results in a considerable effort. For
example, the required material and tools have to be made available,
computerized numerical control (CNC) programs need to be completed
upon start of production, etc.

In the work of \citet{lin:1994:article}, only operations with earlier
starting times are considered in the proposed stability measure.
This implies that delays do not affect the schedules stability while
earlier starting times do, see Expression~(\ref{eqn:stability:2}).

\begin{equation}
\label{eqn:stability:2} stab(x,x') = \sum \max (0, s_{jk} - s_{jk}')
\end{equation}

To some extent, the approach of \citet{lin:1994:article} can be seen as an
answer to the criticism mentioned above with regard to the work of
\citet{wu:1993:article}. The
stability measure is therefore suitable for manufacturing
environments where only earlier executions of operations present a
relevant change to the production plan.

The approaches of
\citet{wu:1993:article} and \citet{lin:1994:article} are combined in the
work presented by \citet{rangsaritratsamee:2004:article} into the definition
of an overall cost function. As a result, a more general way of
measuring the stability of schedules is derived, making it possible
to individually discriminate between the effect of earlier and later
scheduled operations.

As opposed to starting times of operations, \citet{cowling:2002:article} propose a combination of start and completion times of the entire job, leading to an overall cost function. In a later work of \citet{cowling:2004:article}, only the completion times $C_{j}$ of jobs are relevant for the proposed stability measure. As in this stability measure it is not the operations that are considered but the completion of the job as a whole, this concept differs significantly from other approaches of measuring the stability of schedules. Its use can be seen in manufacturing environments where changes of the completion of the jobs are more important than the operations themselves.

The approach of \citet{watatani:1992:inproceedings}, later also used by
\citet{iima:2005:inproceedings}, is based on the sequences of the
operations. Here, schedules $x$ and $x'$ are considered to be
different if the sequences of the operations differ. This is the
case for operations $O_{jk}, O_{lo}$ for which $s_{jk} < s_{lo}
\wedge s_{jk}' > s_{lo}'$. This measure is useful when changing the
sequence of operations is difficult with respect to the organization
of the manufacturing environment and therefore results in a
necessary organizational effort. An example of such a production
situation would be a flow-oriented manufacturing environment where
jobs are transported in a fixed sequence, e.\,g.\ by means of a
belt. Changes to the sequence through jobs overtaking others impose
problems here.

Stability measures are implemented in rescheduling strategies either
in a predictive-reactive way, where the problem is treated as a
multi-objective optimization problem maximizing both utility and
stability, or in concepts of robust scheduling, where they are used
to compute a quantitative measure for a schedule robustness with
respect to changes of the manufacturing situation.

\paragraph{Critical analysis}\hspace{0cm}\\
Stability of schedules is always measured by comparison of an
initial production schedule $x$ with a revised schedule $x'$.
Various characteristics of the schedules $x,x'$ may differ from each
other. In particular, the following aspects have to be considered to
be the basis for a further analysis.

1. Different starting times of operations, along with different completion times of the jobs.\\
2. Different operation sequences.\\
3. Different machine assignments of operations.

Consequently, approaches measuring the stability of schedules are
based on either one characteristic or combine several
characteristics to an overall stability measure.

Stability measures express the difference in a quantitative way,
given an overall idea of how \lq{}different\rq{} schedules are.
Existing approaches define stability with respect to starting time
changes of operations
\citep{wu:1993:article,lin:1994:article,rangsaritratsamee:2004:article,cowling:2002:article,cowling:2004:article}
or changes of operation sequences \citep{iima:2005:inproceedings}. A
distinction between earlier and later starting operations is
sometimes made \citep{lin:1994:article}. This expressed the fact that
on one hand an earlier execution is critical in the sense of
schedule stability, while on the other hand a later execution does
not lead to problems for the organization.

The difference between schedules is of particular importance as
changes of the production planning have an organizational impact,
e.\,g.\ on required material, suppliers, etc. A stability measure of
schedules therefore should express the impact of the changes on the
organization.

Reviewing existing approaches of schedule stability, one can observe
that they do not take into consideration the influence of the point
in time when the changes occur. While the absolute deviation of the
starting times from the initially scheduled $s_{jk}$ plays a role,
it is not further analyzed whether the changes appear close to the
actual rescheduling moment or at the end of the planning horizon. In
many real world applications of scheduling, immediate changes will
however have a different impact compared to changes happening in
e.\,g.\ several weeks. The reason behind this can be seen in the
time needed to implement the changes made to the production
schedule. Communication with suppliers and customers requires time.

\section{Proposition of a novel approach}
\paragraph{A generalized instability measure}\hspace{0cm}\\
To overcome the limitations described in the previous section, we
propose a novel approach measuring the \emph{instability} of
schedules. A measure for instability is defined in such a way that
larger values refer to a larger instability. This is in contrast to
approaches found in the literature, which wrongly denote the
measures as measures of stability, although the computed values
increase with increasing instability.

The concept is based on starting times of operations, and measures instability of schedules analyzing two aspects:\\
1. The size of the deviation of the starting times.\\
2. The point of time, when the deviation occurs.

The first aspect, the size of the deviation of the starting times,
can easily be measured by comparing the starting times $s_{jk}$ and
$s_{jk}'$ of the operations. Expression~(\ref{eqn:delta:sjk})
computes their absolute difference, comparing the initial schedule
$x$ and the revised one $x'$.

\begin{equation}
\label{eqn:delta:sjk} \Delta_{jk} = |s_{jk}' - s_{jk}|
\end{equation}

The second aspect, the point of time when the deviation occurs, can
be determined as follows. The exact time when a schedule has to be
modified, is denoted with $t_{0}$. The relative impact of starting
time changes may be expressed depending on how close either the
initially planned start $s_{jk}$ or the revised start $s_{jk}'$ of
the operation $O_{jk}$ is scheduled.
Expression~(\ref{eqn:sjk:closeness}) measures the closeness of the
operation $O_{jk}$ to the start of the planning horizon.

\begin{equation}
\label{eqn:sjk:closeness} dist_{jk} = \min ( s_{jk} - t_{0}, s_{jk}' -
t_{0} ) = \min (s_{jk}, s_{jk}') - t_{0}
\end{equation}

The impact of the closeness $dist_{jk}$ may then be expressed as a
(monotonic) decreasing function $imp (dist_{jk})$. Operations $O_{jk}$ being close to the start of the current planning horizon, and thus having a small $dist_{jk}$, receive a high impact value $imp (dist_{jk})$, while the impact decreases with increasing distance to the moment of rescheduling.\\
A possible way of computing the impact $imp (dist_{jk})$ is given in
Expression~(\ref{eqn:imp}).

\begin{equation}
\label{eqn:imp} imp (dist_{jk}) = I^{dist_{jk}}
\end{equation}

\paragraph{On obtaining statements from the decision maker}\hspace{0cm}\\
The parameter $I$ has to be chosen such that it reflects the length
of the planning horizon. The actual value of the parameter $I$ could
be obtained from the decision maker, by considering how the relative
impact of an operation at the end of the scheduling horizon relates
to the impact at the very beginning, expressed by parameter $pc$.
For example, the decision maker is enabled to express that changes
at the end of the planning horizon have an impact of 30\%
of changes at the beginning of the horizon, therefore $pc
= 0.3$.

With the percentage at the end of the planning horizon $pc$ and the
length of the planning horizon $T$, $I$ may be computed as given in Expression~(\ref{eqn:computing:I}). A value of  $I < 1$ has been chosen to show a decreasing impact over time.

\begin{equation}
\label{eqn:computing:I}I = \sqrt[T]{pc}
\end{equation}

In practice, it might be difficult for the decision maker to state
the exact percentage of the impact at the end of the planning
horizon. An alternative way for obtaining $I$ could be to refer to another period with which the planner is more familiar. From a practical planner's perspective, such a period could be a (working) week, or a similar time span being used as a time pattern in planning. The decision maker is
then asked about the decrease of the impact within this period,
denoted by $dec$. With a decrease of e.\,g.\ 20\%, $dec = 0.2$ and
the length of a week of five working days, $I$ is computed as given in Expression~(\ref{eqn:computing:I2}).

\begin{equation}
\label{eqn:computing:I2}I = \sqrt[5]{1 - dec}
\end{equation}

In comparison to Expression~(\ref{eqn:computing:I}), the alternative way of computing $I$ in Expression~(\ref{eqn:computing:I2}) only differs with respect to the statement of the decision maker. While in~(\ref{eqn:computing:I}) the decision maker has to state the impact at the end of the planning horizon directly, it is indirectly computed in (\ref{eqn:computing:I2}) using the decrease $dec$ over an arbitrary period.\\
It should be noticed that the impact asymptotically approaches zero
without ever reaching it. Therefore, the impact of starting time
changes decreases over time but always stays positive.

\paragraph{Overall formula}\hspace{0cm}\\
A computation of the total impact of the starting time changes and
therefore the instability $instab(x, x')$ combines the relative
importance of the impact with the size of the change.
Expression~(\ref{eqn:instability}) gives the precise formula.

\begin{equation}
\label{eqn:instability}instab(x,x') = \sum_{j} \sum_{k} imp
(dist_{jk}) \Delta_{jk} = \sum_{j} \sum_{k} \left( I^{\min ( s_{jk},
s_{jk}' ) - t_{0}} \quad | s_{jk}' - s_{jk} | \right)
\end{equation}

In brief this means that combinations of values $s_{jk}, s_{jk}'$
close to $t_{0}$ and big changes of the starting times $| s_{jk} -
s_{jk}' |$ lead to a high impact. On the other hand, small changes
of $s_{jk}$ to $s_{jk}'$ and changes occurring towards the end of
the planning horizon do not contribute as much to the impact
measure. It becomes clear that the required effort for computing Expression~(\ref{eqn:instability}) increases with the number of operations. Also, the exponential component adds to the running time of of the formula. For a practical application, e.\,g.\ in a metaheuristic approach employing local search, where numerous evaluations of Expression~(\ref{eqn:instability}) are required, a closer look at the running time behavior of our approach will become necessary.

\section{\label{sec:discussion:conclusions}Discussion and conclusions}
An approach to measure the instability of schedules in a dynamic
manufacturing environment has been presented. The concept integrates
two aspects into a single measure, namely the difference of starting
times and the significance of the changes depending on when
they occur. As a result, the concept generalizes approaches known
from literature by providing the possibility of assigning a relative
importance of the changes.

The proposed instability measure is more general, yet it includes
the special case of \citet{wu:1993:article} for a chosen value of $I = 1$. In any other case however, the determination of an appropriate $I$ requires
considerable more information compared to existing approaches. Small values of $I$ lead to a fast decrease of the impact, while large values tend to discriminate less between the changes.

Given the possibility to collect information about the relative importance of the changes as described, the approach may reflect the actual practical situation of dynamic scheduling in manufacturing environments more closely. Future research will therefore focus on the evaluation of the applicability of the approach in dynamic machine scheduling situations. Such an attempt could be experimentally-driven, investigating the obtained results when making use of the proposed measure, e.\,g.\ for an exemplary dynamic job-shop- or flow-shop scheduling environment. Different rescheduling factors should be examined, and the impact of well-chosen rescheduling reasons (disturbances) on the obtained schedule should be studied. This also implies an experimental comparison of the proposed measure to approaches known from the literature.


\begin{thebibliography}{17}
\providecommand{\natexlab}[1]{#1}
\providecommand{\url}[1]{\texttt{#1}}
\expandafter\ifx\csname urlstyle\endcsname\relax
  \providecommand{\doi}[1]{doi: #1}\else
  \providecommand{\doi}{doi: \begingroup \urlstyle{rm}\Url}\fi

\bibitem[Abumaizar and Svestka(1997)]{abumaizar:1997:article}
R.J. Abumaizar and J.~A. Svestka.
\newblock Rescheduling job shops under disruptions.
\newblock \emph{International Journal of Production Research}, 35\penalty0
  (7):\penalty0 2065--2082, 1997.

\bibitem[Aytug et~al.(2005)Aytug, Lawley, McKay, Mohan, and
  Uzsoy]{aytug:2005:article}
Haldun Aytug, Mark~A. Lawley, Kenneth McKay, Shantha Mohan, and Reha Uzsoy.
\newblock Executing production schedules in the face of uncertainties: A review
  and some future directions.
\newblock \emph{European Journal of Operational Research}, 161:\penalty0
  86--110, 2005.

\bibitem[Church and Uzsoy(1992)]{church:1992:article}
L.K. Church and R.~Uzsoy.
\newblock Analysis of periodic and event-driven rescheduling policies in
  dynamic shops.
\newblock \emph{International Journal of Computer Integrated Manufacturing},
  5:\penalty0 153--163, 1992.

\bibitem[Cowling et~al.(2004)Cowling, Ouelhadj, and
  Petrovic]{cowling:2004:article}
P.~I. Cowling, D.~Ouelhadj, and S.~Petrovic.
\newblock Dynamic scheduling of steel casting and milling using multi-agents.
\newblock \emph{Production Planning \& Control}, 15\penalty0 (2):\penalty0
  178--188, 2004.

\bibitem[Cowling and Johansson(2002)]{cowling:2002:article}
Peter Cowling and Marcus Johansson.
\newblock Using real time information for effective dynamic scheduling.
\newblock \emph{European Journal of Operational Research}, 139:\penalty0
  230--244, 2002.

\bibitem[Dutta(1990)]{dutta:1990:article}
A.~Dutta.
\newblock Reacting to scheduling exceptions in {FMS} environments.
\newblock \emph{IIE Transactions}, 22\penalty0 (4):\penalty0 300--314, 1990.

\bibitem[Fang and Xi(1997)]{fang:1997:article}
J.~Fang and Y.~Xi.
\newblock A rolling horizon job shop rescheduling strategy in the dynamic
  environment.
\newblock \emph{International Journal of Advanced Manufacturing Technology},
  13:\penalty0 227--232, 1997.

\bibitem[Haupt(1989)]{haupt:1989:article}
R.~Haupt.
\newblock A survey of priority rule-based scheduling.
\newblock \emph{Operations Research Spektrum}, 11\penalty0 (1):\penalty0 3--16,
  1989.

\bibitem[Iima(2005)]{iima:2005:inproceedings}
Hitoshi Iima.
\newblock Proposition of selection operation in a genetic algorithm for a job
  shop rescheduling problem.
\newblock In Carlos~A. {Coello Coello}, Arturo {Hern\'{a}ndez Aguirre}, and
  Eckart Zitzler, editors, \emph{Evolutionary Multi-Criterion Optimization},
  Lecture Notes in Computer Science 3410, pages 721--735, Berlin, Heidelberg,
  New York, 2005. Springer Verlag.

\bibitem[Jain and Elmaraghy(1997)]{jain:1997:article}
A.K. Jain and H.A. Elmaraghy.
\newblock Production scheduling/rescheduling in flexible manufacturing.
\newblock \emph{International Journal of Production Research}, 35\penalty0
  (1):\penalty0 289--309, 1997.

\bibitem[Kouvelis et~al.(2000)Kouvelis, Daniels, and
  Vairaktarakis]{kouvelis:2000:article}
P.~Kouvelis, R.~L. Daniels, and G.~Vairaktarakis.
\newblock Robust scheduling of a two-machine flow shop with uncertain
  processing times.
\newblock \emph{IIE Transactions on Scheduling and Logistics}, 32:\penalty0
  421--432, 2000.

\bibitem[Lin et~al.(1994)Lin, Krajewski, Leong, and Benton]{lin:1994:article}
N.~P. Lin, L.~Krajewski, G.~K. Leong, and W.~C. Benton.
\newblock The effects of environmental factors on the design of master
  production scheduling systems.
\newblock \emph{Journal of Operations Management}, 11:\penalty0 367--384, 1994.

\bibitem[Mehta and Uzsoy(1999)]{mehta:1999:article}
Sanjay~V. Mehta and Reha Uzsoy.
\newblock Predictable scheduling of a single machine subject to breakdowns.
\newblock \emph{International Journal of Computer Integrated Manufacturing},
  12\penalty0 (1):\penalty0 15--38, 1999.

\bibitem[Rangsaritratsanee et~al.(2004)Rangsaritratsanee, Jr., and
  Kurz]{rangsaritratsamee:2004:article}
Duedee Rangsaritratsanee, William G.~Ferrell Jr., and Mary~Beth Kurz.
\newblock Dynamic rescheduling that simultaneously considers efficiency and
  stability.
\newblock \emph{Computers \& Industrial Engineering}, 46:\penalty0 1--15, 2004.

\bibitem[T'kindt and Billaut(2002)]{tkindt:2002:book}
Vincent T'kindt and Jean-Charles Billaut.
\newblock \emph{Multicriteria Scheduling: Theory, Models and Algorithms}.
\newblock Springer Verlag, Berlin, Heidelberg, New York, 2002.

\bibitem[Watatani and Fujii(1992)]{watatani:1992:inproceedings}
Y.~Watatani and S.~Fujii.
\newblock A study on rescheduling policy in production systems.
\newblock In \emph{Proceedings of the JAPAN/USA Symposium on Flexible
  Automation, ASME}, pages 1147--1150, 1992.

\bibitem[Wu et~al.(1993)Wu, Storer, and Chang]{wu:1993:article}
S.~David Wu, Robert~H. Storer, and Pei-Chann Chang.
\newblock One-machine rescheduling heuristics with efficiency and stability as
  criteria.
\newblock \emph{Computers \& Operations Research}, 20\penalty0 (1):\penalty0
  1--14, 1993.

\end{thebibliography}
\end{document}